\documentclass{article}

\usepackage{PRIMEarxiv}

\usepackage[utf8]{inputenc} 
\usepackage[T1]{fontenc}    
\usepackage{hyperref}       
\usepackage{url}            
\usepackage{booktabs}       
\usepackage{amsfonts}       
\usepackage{nicefrac}       
\usepackage{microtype}      
\usepackage{lipsum}
\usepackage{fancyhdr}       
\usepackage{graphicx}       
\usepackage{orcidlink}
\graphicspath{{./images/}}     
\usepackage{subfig}
\usepackage{float}

\title{Image captioning for Brazilian Portuguese using GRIT model

}

\author{
  Rafael Silva de Alencar \\
  Alana AI Research \\
  São Paulo, Brazil\\
  \texttt{rafael.silva@alana.ai} \\
   \And
  William Alberto Cruz Castañeda
  \orcidlink{0000-0002-9803-1387}\\
  Alana AI Research \\
  São Paulo, Brazil \\
  \texttt{william.cruz@alana.ai} \\
  \And
  Marcellus Amadeus \orcidlink{0009-0002-7777-2562}\\
  Alana AI Research \\
  São Paulo, Brazil\\
  \texttt{marcellus@alana.ai} \\
}

\begin{document}
\maketitle

\begin{abstract}
This work presents the early development of a model of image captioning for the Brazilian Portuguese language. We used the GRIT (Grid - and Region-based Image captioning Transformer) model to accomplish this work. GRIT is a Transformer-only neural architecture that effectively utilizes two visual features to generate better captions. The GRIT method emerged as a proposal to be a more efficient way to generate image captioning. In this work, we adapt the GRIT model to be trained in a Brazilian Portuguese dataset to have an image captioning method for the Brazilian Portuguese Language.

\end{abstract}

\keywords{Image Captioning \and Brazilian Portuguese Captioning \and Grid Features \and Region Features}

\section{Introduction}

Image captioning is a new trend in the Machine Learning (ML) field heavily studied in the past few years. The goal is to generate a semantically understandable text given an image. Several studies have unique approaches to solving this problem with Transformers as a fundamental basis. Exists two primary methods for the Imagine captioning problem: grid features and region features. Grid features are local image features extracted at the regular grid points, often obtained directly from a higher layer feature map(s) of CNNs/ViTs. Region features are a set of local image features of the regions (i.e., bounding boxes) detected by an object detector\cite{nguyen2022grit}.
In this study, were integrated the two methods to build a better model for image captioning. With the integration of these two methods, the model will provide a better representation of input images since they are complementary to each other\cite{nguyen2022grit}.

These components form a Transformer-only neural architecture, dubbed GRIT(Grid- and Region-based Image captioning Transformer). The experimental results from the study\cite{nguyen2022grit}  show that GRIT has established a new state-of-the-art on standard image captioning. In this work, we used a Brazilian Portuguese-language translated version of the COCO dataset\cite{lin2014microsoft}. Besides the dataset, the model also requires a vocabulary file, which is a set of words present in the corpus of the dataset. The experiment was conducted with the COCO dataset in Brazilian Portuguese to obtain a model that comprehends other language than English.

\section{Method}
The GRIT (Grid- and Region-based Image captioning Transformer) consists of two parts, one for extracting the dual visual features from an input image and the other for generating a caption sentence from the extracted features \cite{nguyen2022grit}. The model architecture is described below in Figure \ref{fig:fig1}.

\begin{figure}[H]
    \centering
    \includegraphics[width=16cm]{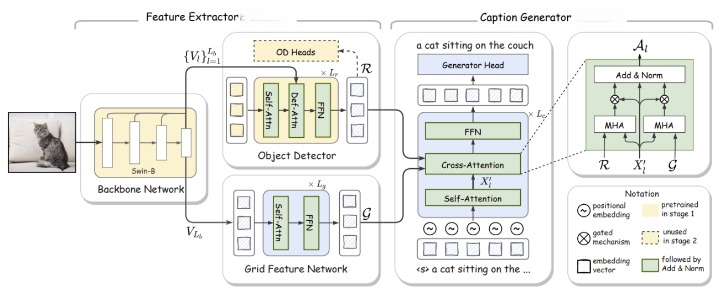}
    \caption{GRIT model architecture\cite{nguyen2022grit}}
    \label{fig:fig1}
\end{figure}

\subsection{Extraction of Visual Features from Images}

In the first step in the GRIT model, a Swin Transformer applies initial visual features from the input image. This first step is mentioned in the paper as Backbone Network for Extracting Initial Features\cite{nguyen2022grit}. The second step is generating Region Features. It employs a transformer-based decoder framework called DETR, instead of CNN-based detectors, such as Faster R-CNN used by SOTA image captioning models\cite{nguyen2022grit}. Afterward, the Grid Feature Network receives the last one of the multi-scale feature maps from the Swin Transformer backbone, i.e.,    $V_{L_b}$ $\in$  $\mathbf{R}^{d \times d_{L_b} }$, where $M = H$/64 $\times$ W/64\cite{nguyen2022grit}.

\subsection{Caption Generation Using Dual Visual Features}

The caption generator receives the two types of visual features, the region features and the grid features, as inputs. It generates a caption sentence in an autoregressive manner. That is, receiving the sequence of predicted words (rigorously their embeddings) at time t $-$ 1. It predicts the next word at time t. We employ the sinusoidal positional embedding of the time step; we add it to the word embedding to obtain the input $x_0^t$ $\in$ $\mathbb{R}^d$ at $t$\cite{nguyen2022grit}. The caption generator consists of a stack of $L_c$ identical layers. The initial layer receives the sequence of predicted words and the output from the last layer is input to a linear layer whose output dimension equals the vocabulary size to predict the next word\cite{nguyen2022grit}.

\section{Experimental Setup}

In this section, we describe the experimental setup used for training the Image-To-Text (ITT) GRIT model\cite{nguyen2022grit} for the Brazilian Portuguese language. It also informed the hardware and software infrastructure, specific settings, the benchmark used, metrics, and dataset. The experiment was conducted entirely with the same steps taken in the original GRIT model in English, which are the extraction of visual features from images and Caption Generation using dual visual features. It was not required to make structural changes in the original model to produce similar results for the Brazilian Portuguese version.  

\subsection{Hardware and Softwares}
Implementing the original ITT GRIT model requires two machines with a GPU A100 80GB. The experiment applied one epoch compared with the original GRIT \cite{nguyen2022grit}, which has ten epochs, and the whole training process took approximately seven hours and thirty minutes to complete. All the configuration for the experiment was in a Linux Ubuntu Operating System version 20.0.

\subsection{Datasets}
Showing the similarities and differences between the English GRIT and Brazilian Portuguese GRIT models, we describe the datasets used.

\textbf{English Image Captioning} \ The English GRIT model uses the COCO dataset from the original experiment. The dataset contains 123287 images. Each image has five different annotation captions. The experiment uses 113287 images in the training process and 5000 for validation and testing. 

\textbf{Portuguese Image Captioning} \ The Brazilian Portuguese GRIT model uses a translated version of the COCO dataset. This dataset contains the same images as the original COCO (123,287). Each image has five different annotation captions translated into the Brazilian Portuguese language. The experiment uses 113,287 images in the training process and 5,000 for validation and testing.

An important thing to point out is that the training of the dataset also involves the use of a vocabulary, which is essentially a list of tokenized words present in the corpus of the dataset. The details of how it works in the training process are yet to be known; the work \cite{nguyen2022grit} does not show details of implementation left to programmers, the interpretation of the scripts, and the study of how the vocabulary is involved in the process.

ITT English GRIT model experiments implement a vocabulary containing 10201 words. In our attempt to train in the translated version of the GRIT model, we took the original vocabulary file and performed a literal translation of the file. So, in this first attempt to produce an ITT Brazilian Portuguese model, we use a vocabulary, which is a literal translation of the original.

\subsection{Evaluation Metrics}
To evaluate the quality of the model it was employed the metrics BLEU@N\cite{papineni2002bleu}, METEOR\cite{banerjee2005meteor}, ROUGE-L\cite{lin2004rouge}, CIDEr\cite{vedantam2015cider}.

\textbf{BLEU} (BiLingual Evaluation Understudy) is an automatic evaluation metric of machine-translated text. The BLEU score is a number between zero and one that compares the similarity of machine-translated text to a set of high-quality reference translations. The value 0 means that the machine translation output does not match the reference translation (low quality). Value 1 means perfect match with reference translations (high quality)\cite{Google}. The values in table \ref{tab:table_1} are on a scale between 0 e 1, where 0 means 0$\%$ and 1 means 100$\%$.

\begin{table}[H]
\centering
\caption{Possible outcomes of the BLEU metric\cite{Google}.}
\begin{tabular}{ll}
BLEU Score                            & Interpretation                                                                      \\ \hline
\multicolumn{1}{|l|}{\textless 10}    & \multicolumn{1}{l|}{Practically useless}                                            \\ \hline
\multicolumn{1}{|l|}{10 - 19}         & \multicolumn{1}{l|}{Difficult to understand the meaning}                            \\ \hline
\multicolumn{1}{|l|}{20 - 29}         & \multicolumn{1}{l|}{The meaning is clear, but there are serious grammatical errors} \\ \hline
\multicolumn{1}{|l|}{30 - 40}         & \multicolumn{1}{l|}{Can be understood as good translations}                         \\ \hline
\multicolumn{1}{|l|}{40 - 50}         & \multicolumn{1}{l|}{High-quality translations}                                      \\ \hline
\multicolumn{1}{|l|}{50 - 60}         & \multicolumn{1}{l|}{Very high quality, adequate and fluent translations}            \\ \hline
\multicolumn{1}{|l|}{\textgreater 60} & \multicolumn{1}{l|}{In general, higher than human quality}                          \\ \hline
\end{tabular}
\label{tab:table_1}
\end{table}

\textbf{METEOR}, is an automatic metric for machine translation evaluation based on a generalized concept of unigram matching between machine-produced translation and human-produced reference translations. Unigrams can be matched based on their surface forms, stemmed forms, and meanings. Once all generalized unigram matches between the two strings have been found, METEOR computes a score for this matching using a combination of unigram-precision, unigram-recall, and a measure of fragmentation designed to capture how well-ordered the matched words in the machine translation are about the reference\cite{banerjee2005meteor}. METEOR gets an R correlation value of 0.347 with human evaluation on the Arabic data and 0.331 on the Chinese data.

\textbf{CIDEr} A Consensus-based Image Description Evaluation metric for evaluating the quality of generated textual descriptions of images. The CIDEr metric measures the similarity between a generated caption and the reference captions. Based on the concept of consensus: the idea that good captions should not only be similar to the reference captions in terms of word choice and grammar but also in terms of meaning and content \cite{vedantam2015cider}.

\textbf{ROUGE}, or Recall-Oriented Understudy for Gisting Evaluation, is a set of metrics and a software package for evaluating automatic summarization and machine translation software in natural language processing. The metrics compare an automatically produced summary or translation against a reference or a set of references (human-produced) summary or translation. ROUGE metrics range between 0 and 1, with higher scores indicating higher similarity between the automatically produced summary and the reference\cite{lin2004rouge}.

\section{Results}

\begin{figure}[H]
\centering
\parbox{6cm}{
\includegraphics[width=6cm]{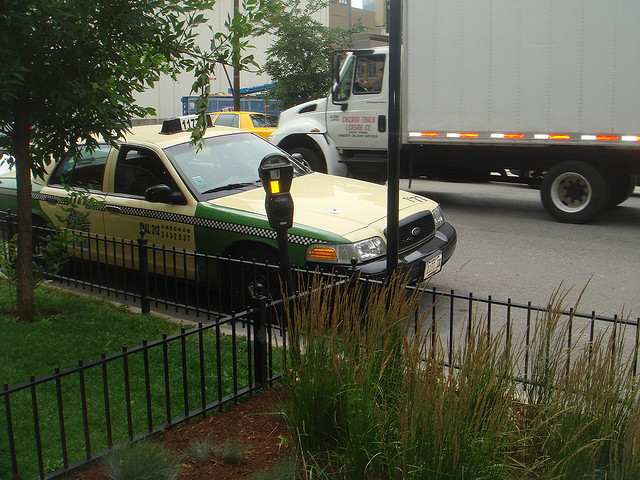}
\caption{"um carro Estacionado em o lado de um rua"}
\label{fig:fig2}}
\qquad
\begin{minipage}{6cm}
\includegraphics[width=6cm]{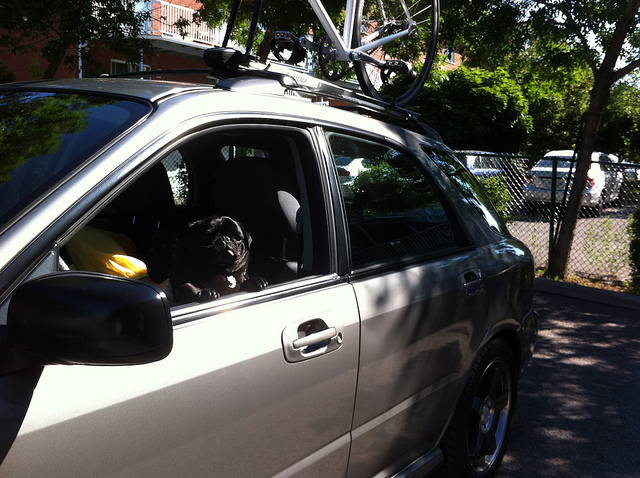}
\caption{"um cão Isso É sentado em o Voltar de um carro"}
\label{fig:fig3}
\end{minipage}
\end{figure}

\begin{figure}[H]
    \centering
    \includegraphics[width=6cm]{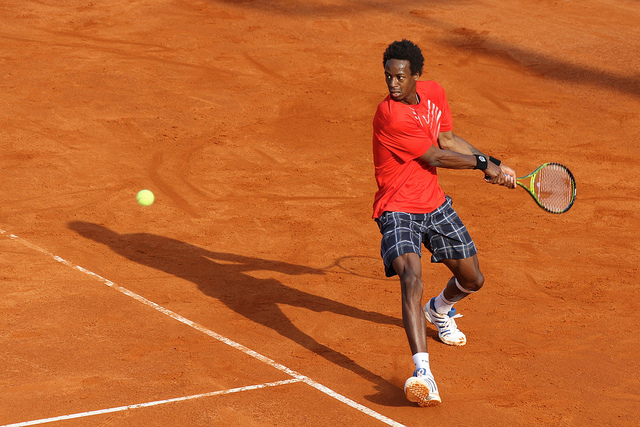}
    \caption{"um homem Jogar tênis em um tênis tribunal"}
    \label{fig:fig4}
\end{figure}

This section shows the model results after obtaining the checkpoints from the training steps. With the checkpoint, we implement a script that generates the captions in Brazilian Portuguese for a given image. The model generates captions that are still missing a better semantic quality. Legibly in Brazilian Portuguese and relates to the image associated with it. The experiment initializes with the same data settings as the original experiment. In the first instance, the goal was to test the reproducibility of the model and produce a first attempt at a Brazilian Portuguese version of the model. It was possible to initiate and finish the training process given one epoch. Through the examples shown below, we can observe the resulting process of the experiment. We show captions generated by an image as input. In the images \ref{fig:fig2}, \ref{fig:fig3} and \ref{fig:fig4} we have a fairly accurate description through the caption, however, there are some issues with literal translation and semantics.  

Table \ref{tab:table_2} shows the metrics results of the Brazilian Portuguese version of the GRIT model. Compared with the original English GRIT model, the results are different but not extremely far, which means that even running the model with one epoch achieves similar results. 

\begin{table}[H]
\centering
\caption{Brazilian Portuguese GRIT model result metrics}
\begin{tabular}{|l|c|c|}
\hline
Metrics & Portuguese model & English model \cite{nguyen2022grit} \\ \hline
BLEU    & 0.758            & 0.842         \\ \hline
METEOR  & 0.268            & 0.306         \\ \hline
ROUGE   & 0.557            & 0.607         \\ \hline
CIDEr   & 1.100            & 1.442         \\ \hline
\end{tabular}
\label{tab:table_2}
\end{table}

\section{Conclusion}

The main objective of this work is to propose a Transformer-based architecture for image captioning named GRIT for Brazilian Portuguese. The idea is to integrate the region features and the grid features extracted from an input image to extract richer visual information from input images. The experimental results validated our approach, showing that GRIT outperforms all published methods by a large margin in inference accuracy and speed. In our version of the ITT GRIT model, we used a dataset in Brazilian Portuguese, which is a translation from the original COCO dataset. The experiment uses the same steps taken in the original GRIT model. 

As a future work, the main goal is to produce another prototype of the ITT GRIT model for the Brazilian Portuguese language without a vocabulary file. As a reference, we are working on a different branch of the GRIT model (vicap branch)\cite{nguyen2022grit}. This branch is specifically for fine-tuning other datasets besides the COCO.

\section*{Acknowledgments}
We acknowledge the support of the National Council for Scientific and Technological Development (CNPq), and Alana AI for funding the research.

\section*{Funding}
This research is supported by the National Council for Scientific and Technological Development (CNPq/MCTI/SEMPI Number 021/2021 RHAE - 384217/2023-0), Ministry of Science, Technology and Innovation, and Alana AI. 

\section*{Contributions}
RA conducted the experiments. WC and MS provided oversight for the research. All authors contributed to the paper's composition, with MS and WC contributing to the study's conception. The final manuscript was reviewed and approved by all authors.

\section*{Interests}
The authors declare that they have no competing interests.

\section*{Materials}
The datasets generated and/or analyzed during the current study are available on request.

\bibliographystyle{unsrt}  
\bibliography{references}

\end{document}